\documentclass[10pt,twocolumn,letterpaper]{article}

\usepackage{wacv}
\usepackage{times}
\usepackage{epsfig}
\usepackage{graphicx}
\usepackage{amsmath}
\usepackage{amssymb}
\usepackage[skip=2pt,font=small]{caption}
\usepackage{amsmath}
\usepackage{mathtools}
\usepackage{tabularx,booktabs}
\usepackage{graphicx}
\usepackage{array}
\usepackage[utf8]{inputenc}
\usepackage{tabu}
\usepackage{tabularx}
\usepackage{booktabs}
\usepackage[english]{babel}
\usepackage{csquotes}
\usepackage[skip=2pt,font=small]{caption}
\usepackage{footnote}
\usepackage{dblfloatfix}
\usepackage{amssymb}
\usepackage{multirow}
\usepackage{afterpage}

\def\wacvPaperID{1322} 

\wacvfinalcopy 

\ifwacvfinal
\def\assignedStartPage{9876} 
\fi

\ifwacvfinal
\usepackage[breaklinks=true,bookmarks=false]{hyperref}
\else
\usepackage[pagebackref=true,breaklinks=true,colorlinks,bookmarks=false]{hyperref}
\fi

\ifwacvfinal
\else
\fi

\begin{document}

\title{
SASRA: Semantically-aware Spatio-temporal Reasoning Agent \\
for Vision-and-Language Navigation in Continuous Environments}
\author{
        Muhammad Zubair Irshad\textsuperscript{*} \ \ \  \ 
        Niluthpol Chowdhury Mithun\textsuperscript{\dag} \ \ \ \
        Zachary Seymour\textsuperscript{\dag} \ \
        Han-Pang Chiu\textsuperscript{\dag} \\
        Supun Samarasekera\textsuperscript{\dag} \ \ \ \
        Rakesh Kumar\textsuperscript{\dag} \\
         \textsuperscript{*}Georgia Institute of Technology \ \ \ \ \ \textsuperscript{\dag}SRI International \ \ \ \ \ \
}

\newcommand{\ssymbol}[1]{^{\@fnsymbol{#1}}}
\twocolumn[{%
\renewcommand\twocolumn[1][]{#1}%
\maketitle
\begin{center}
    \includegraphics[width=0.98\textwidth]{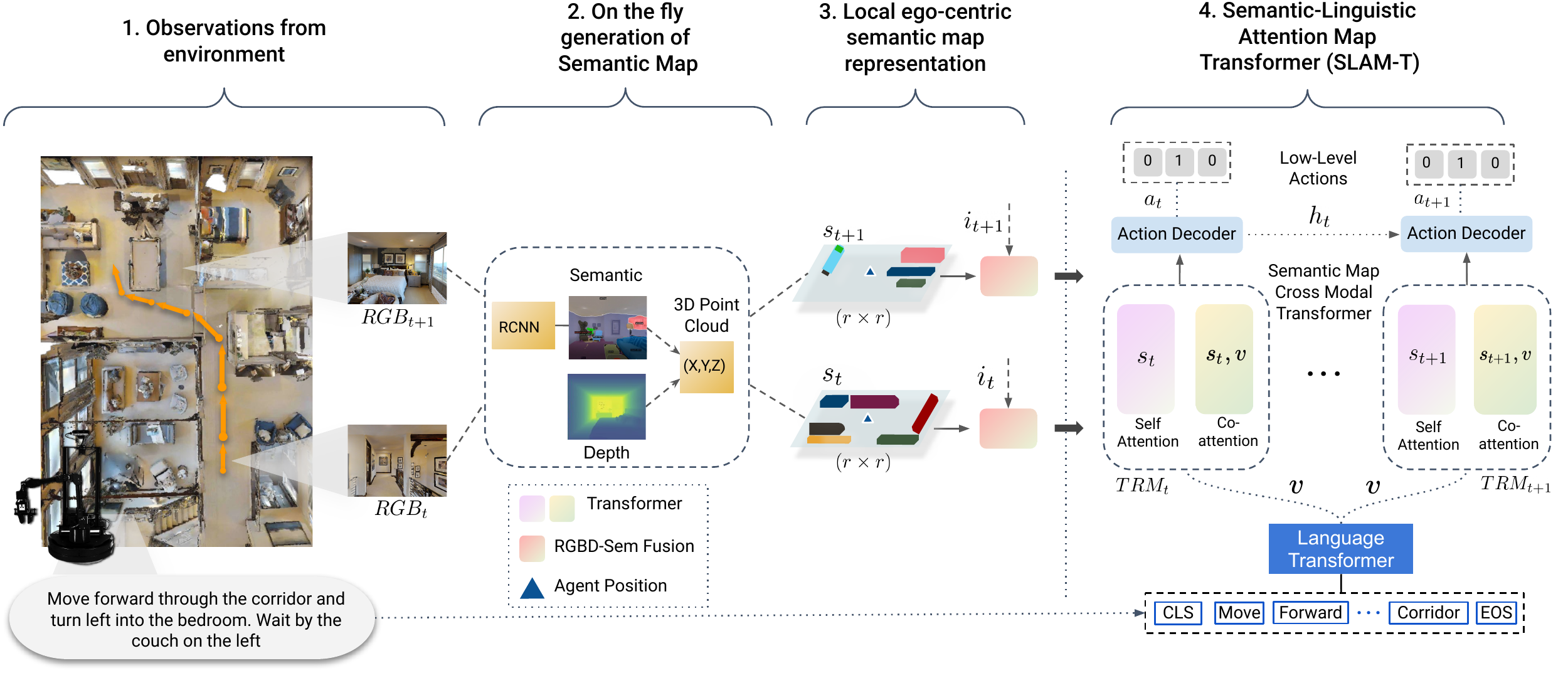}
    \captionof{figure}{\textbf{Overview:} Vision and Language Navigation task and our proposed SASRA agent. Our main novelty lies in employing a hybrid transformer-recurrence model for VLN by utilising a cross-modal Semantic-Linguistic Attention Map Transformer (SLAM-T). At each time-step $t$, the agent generates a local semantic map from visual observations. The agent consistently reasons with the environment in both  spatial (cross modal attention between semantic map and language) and temporal (preserving previous states information through time) domains to decode the low-level actions ($a_{t}$) at each time-step.}
  \label{overview}
  \vspace{0.1cm}
\end{center}%
}]
{
 \renewcommand{\thefootnote}%
    {\fnsymbol{footnote}}
 \footnotetext[1]{Work done while MZI was an intern at SRI International}
 \footnotetext[1]{Georgia Institute of Technology {\tt\scriptsize mirshad7@gatech.edu}}
 \footnotetext[2]{SRI International {\tt\scriptsize (niluthpol.mithun, zachary.seymour, han-pang.chiu, supun.samarasakera, rakesh.kumar)@sri.com}}
}
\thispagestyle{empty}
\pagestyle{empty}

\begin{abstract}

This paper presents a novel approach for the Vision-and-Language Navigation (VLN) task in continuous 3D environments, which requires an autonomous agent to follow natural language instructions in unseen environments. Existing end-to-end learning-based VLN methods struggle at this task as they focus mostly on utilizing raw visual observations and lack the semantic spatio-temporal reasoning capabilities which is crucial in generalizing to new environments. In this regard, we present a hybrid transformer-recurrence model which focuses on combining classical semantic mapping techniques with a learning-based method. Our method creates a temporal semantic memory by building a top-down local ego-centric semantic map and performs cross-modal grounding to align map and language modalities to enable effective learning of VLN policy. Empirical results in a photo-realistic long-horizon simulation environment show that the proposed approach outperforms a variety of state-of-the-art methods and baselines with over $22\%$ relative improvement in SPL in prior unseen environments.

\end{abstract}

\section{INTRODUCTION}

A long-held vision for robotics and artificial intelligence is to create robots that can reason about their environments and perform a task following natural language commands. Recent advances in learning-based approaches~\cite{embodiedqa,johnson2017clevr, Matterport3D} have shown promise in making progress towards solving a complex task referred to as Vision-and-Language Navigation (VLN)~\cite{mattersim, krantz2020navgraph}. VLN task requires an autonomous agent to navigate in a 3D environment without prior access to a map following human-like rich instructions. 

Consider an instruction as shown in Figure \ref{overview}: \enquote{Move forward through the corridor and turn left into the bedroom. Wait by the couch on the left}. In following this sequence of instructions, the agent has to keep track of the preliminary instruction that has been completed as well as keep a record of the seen areas of the environment. This entails the agent to possess an excellent memory. The agent also has to be spatially-aware of its surroundings by matching the visual cues observed at each time-step with relevant parts of the provided instructions. Lastly, the agent has to take low-level control decisions which have long term consequences for instances bumping into obstacles in future. These challenges require the agent to be both spatially-aware of the current scene and temporally conscious of its environment. 

Recent advances in learning-based approaches have made promising progress towards solving the VLN task~\cite{mattersim, krantz2020navgraph,ma2019theregretful, Wang2019ReinforcedCM, hong2020recurrent, wang2021structured}. However, prior end-to-end learning based VLN systems mostly rely on raw visual observations~\cite{ma2019theregretful, Wang2019ReinforcedCM, DBLP:conf/nips/FriedHCRAMBSKD18, krantz2020navgraph, 10.5555/2969033.2969173, Zhuetal,46942} and memory components such as LSTM~\cite{Matterport3D}. Most of the prior VLN systems are trained assuming a simpler navigation-graph based setting (e.g.,\cite{mattersim, ma2019theregretful, Wang2019ReinforcedCM, DBLP:conf/nips/FriedHCRAMBSKD18, 10.5555/2969033.2969173, Zhuetal,46942}), but the performance of these agents is adversely affected in more complex scenarios such as unconstrained navigation or complex long-horizon VLN problems ~\cite{krantz2020navgraph, ai2thor, irshad2021hierarchical}. In this work, we consider a continuous 3D environment for VLN following~\cite{krantz2020navgraph}. The task is referred to as Vision-and-Language Navigation in Continuous Environments (VLN-CE). The VLN-CE formulation imitate the challenges of real-world much better compared to the navigation-graph based formulation and robotic agents trained on such a simulated setting are likely to transfer well to real-world cases.

Structured memory such as local occupancy maps~\cite{chaplot2020learning, Chen_2019_CVPR, gupta2017cognitive, Savinov2019_EC} and scene semantics~\cite{chaplot2020semantic, DBLP:journals/corr/abs-2007-09841, wang2021structured, maast} have shown to previously improve the performance of learning-based visual navigation agents. However, their relation to Vision-and-Language navigation in continuous environments has not been effectively investigated~\cite{chen2020topological}. We believe structure scene memory is crucial to operate well in VLN-CE setting (as we also show for SASRA experiments in Section \ref{sec:exp}) because a longer horizon task demands elaborate spatial reasoning to more effectively learn a VLN Policy.

Motivated by above, we present a hybrid transformer-recurrence model for VLN in continuous 3D environments which we refer to as \textbf{S}emantically-\textbf{a}ware \textbf{S}patio-temporal  \textbf{R}easoning  \textbf{A}gent (\textbf{SASRA}). Our proposed approach combines classical semantic mapping technique with a learning-based method and equips the agent with the following key abilities: 1) Establish a temporal memory by generating local top-down semantic maps from the first person visual observations (RGB and Depth). 2) Create a spatial relationship between contrasting vision-based mapping and linguistic modalities. Hence, identify the relevant areas in the generated semantic map corresponding to the provided instructions and 3) Preserve the relevant temporal states information through time by combining a transformer-inspired~\cite{NIPS2017_7181} attentive model with a recurrence module; hence make the agent conscious of its environment (spatially scene-aware) as well as its history (temporally conscious).

\vspace{0.1cm}
\noindent The main contributions of the work are summarized below: 

\begin{itemize}

\item To the best of our knowledge, we present the first work on the effective integration of semantic mapping and language in an end-to-end learning methodology to train an agent on the task of Vision-and-Language Navigation in continuous 3D environments.
\item We introduce a novel 2-D cross-modal \textbf{S}emantic-\textbf{L}inguistic \textbf{A}ttention \textbf{M}ap \textbf{T}ransformer (SLAM-T) mechanism, based on Transformers, to extract cross-modal map and language features while focusing on the most relevant areas of structured, local top-down spatial information.

\item We demonstrate a clear improvement over the state-of-the-art methods and baselines in the Vision-and-Language navigation task in VLN-CE dataset with over 20\% improvement in success rate and 22\% improvement in success weighted by path length (SPL) in unseen environments.

\end{itemize}
\section{RELATED WORKS}

\textbf{Vision-and-Language Navigation:} The combination of vision and language have been broadly studied in a variety of tasks including Visual Question Answering~\cite{Das_2018_CVPR, balanced_binary_vqa} and Image Captioning~\cite{Anderson_2018_CVPR, vinyals2014neural}. Previous systems for VLN have focused on utilising raw visual observations with a sequence-to-sequence based approach~\cite{mattersim, deng2020evolving, hao2020towards, ke2019tactical, ma2019theregretful}. Additional progress in VLN systems has been guided through cross-modal matching techniques~\cite{Wang2019ReinforcedCM, DBLP:journals/corr/abs-1905-13358}, auxiliary losses~\cite{ma2019selfmonitoring, Zhuetal}, data-augmentation~\cite{majumdar2020improving, DBLP:conf/nips/FriedHCRAMBSKD18, tan2019learning} and improved training-regimes~\cite{krantz2020navgraph}. Our work, in contrast, focuses on incorporating structured memory in the form of scene semantics and its relation to language for VLN systems. Inspired by recent works on utilising end-to-end Transformers~\cite{NIPS2017_7181} for language understanding, neural machine translation~\cite{devlin-etal-2019-bert} and object detection~\cite{DBLP:journals/corr/abs-2005-12872}, our work focuses on end-to-end Transformers for VLN combining various visual modalities with semantics mapping and language. Our formulation allows both spatial as well as temporal reasoning and is effective towards complex task such as VLN. 
Most of the prior works on VLN operate on a navigation graph setting \cite{mattersim, ma2019theregretful, Wang2019ReinforcedCM, DBLP:conf/nips/FriedHCRAMBSKD18}. However, recent efforts have focused on lifting the assumption of navigation graph by making the VLN problem more challenging and closer to the real-world~\cite{krantz2020navgraph, irshad2021hierarchical,roh2020conditional}. In this work, we focus on Vision-and-Language in Continuous Environments (VLN-CE) proposed in~\cite{krantz2020navgraph}.

\textbf{Scene-aware mapping for Navigation:} 
Local map-building during exploration~\cite{chaplot2020neural, gupta2017cognitive,gupta2017unifying, parisotto2017neural,meng2020scaling, chen2020topological} without prior access to a global map has been an active area of research. Previous works have mainly focused on generating a local occupancy map~\cite{chaplot2020learning, Chen_2019_CVPR, gupta2017cognitive, Savinov2019_EC} for better exploration policies. Although top-down occupancy maps are effective for obstacle-avoidance for visual navigation, they fail to encode the rich semantic cues present in the environment. In contrast, our work accumulates the rich semantic information in the environment in a top-down, egocentric semantic map representation. This formulation allows us to spatially reason with language cues, which is shown to aid performance of our agent. Learning semantic maps have been a focus of recent works~\cite{chaplot2020semantic, DBLP:journals/corr/abs-2007-09841, maast} but it has been studied primarily in the context of visual navigation. Our work, however, combines classical semantic map generation in an attentive transformer-like architecture~\cite{NIPS2017_7181} to focus on and extract cross-modal features between local semantics map and provided instructions for effective Vision-and-Language Navigation. To the best of our knowledge, this intersection of semantic mapping and language has not been a focus of any prior works.

\section{Semantically-aware Spatio-Temporal Reasoning Agent}
\begin{figure}
\vspace{0.2cm}
\begin{center}
\includegraphics[width=0.99\linewidth]{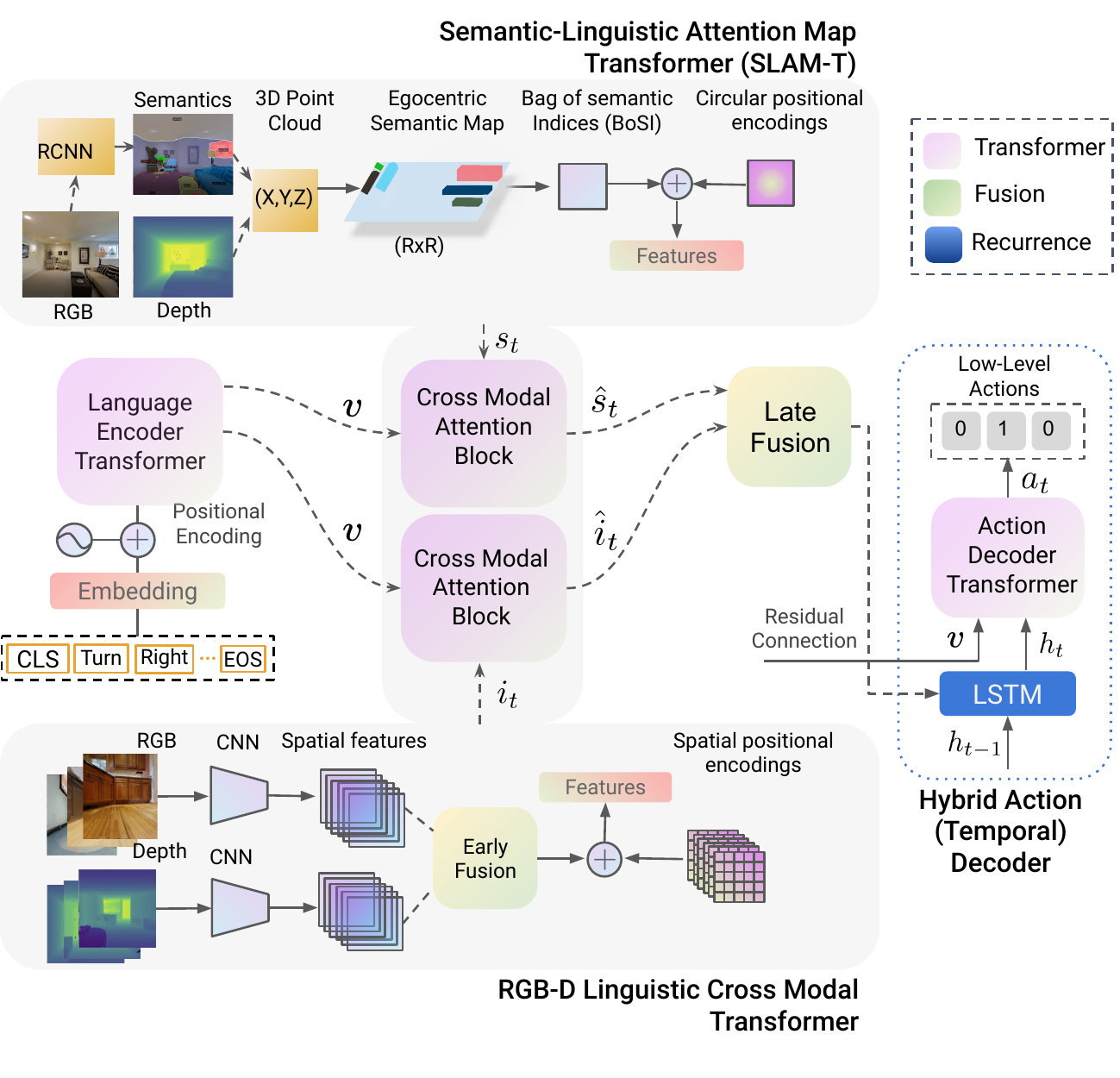}
\captionof{figure}{\textbf{Model Architecture (Detailed model):} Our approach utilises learning-based cross modal attention modules. Semantic-Linguistic Attention Map Transformer (SLAM-T) and RGBD-Linguistic Transformer consistently reason between visual and textual spatial domains. Hybrid Action Decoder captures the temporal dependencies inherent in following a trajectory over time.}
  \label{SASRA2}
\end{center}
\vspace{-0.1cm}
\end{figure}

Consider an autonomous agent $\tilde{\mathcal A}$ in an unknown environment. Our goal is to find the most optimal sequence of actions $a_{t} = \pi(x_{t}, \gamma, q)$ which takes the agent from a starting location to a goal location while staying close to the path defined by instructions ($q$). The agent receives visual observations ($x_{t}$) from the environment at each time-step ($t$). $\gamma$ denotes the learnable parameters of the policy $\pi$. \\
To learn an effective policy ($\pi$), we propose a hybrid transformer-recurrence model to effectively reason with the environment in both spatial and temporal domains. Our approach, as shown in Figure \ref{SASRA2}, combines end-to-end transformer blocks with recurrent modules to effectively match the visual cues with the provided instruction. Our methodology  comprises of three learnable sub-modules: \textbf{S}emantic-\textbf{L}inguistic \textbf{A}ttention \textbf{M}ap \textbf{T}ransformer (SLAM-T), RGBD-Linguistic Cross Modal Transformer and Hybrid Action Decoder. The goal of the SLAM-T module is to generate a local semantic map over time and identify the matching between provided instruction and semantic map using cross attention. Hybrid Action Decoder then utilises the cross attention features obtained from both RGBD-Linguistic and Semantic Map-Linguistic modules to select an action ($a_{t}$) at each time-step. Details of our architecture are summarized as follows:
\subsection{Semantic-Linguistic Attention Map Transformer}

To generate a semantic map from visual observations $x_{t} = \{ r_{t}, d_{t}, g_{t} \}$, we employ a classical three step approach. Here $r_{t}$, $d_{t}$ and $g_{t}$ denote the RGB, Depth and Semantic sensor readings respectively. 

We first project the pixels in depth observations to a 3D point cloud using camera intrinsic parameters. Agent's current pose estimate is used to find the camera extrinsic parameters which are used to transform the point cloud to the world frame. The points are projected to a 2D space and stored as either obstacles based on a height threshold or free space. For each point categorized as an obstacle, we store its semantic class $k$ i.e the value given by the semantic sensor reading $g_{t}$. In essence, each cell in a discrete $2r \times2r$ map $s_{t}$ is represented as a binary vector $s_{t_{xy}}$ $\in$ $\{0, 1\}$,
where $k = 1$ if a semantic class $k$ is present at that location and 0 otherwise.

\begin{figure*}[t!]
\begin{center}
\includegraphics[width=0.88\textwidth]{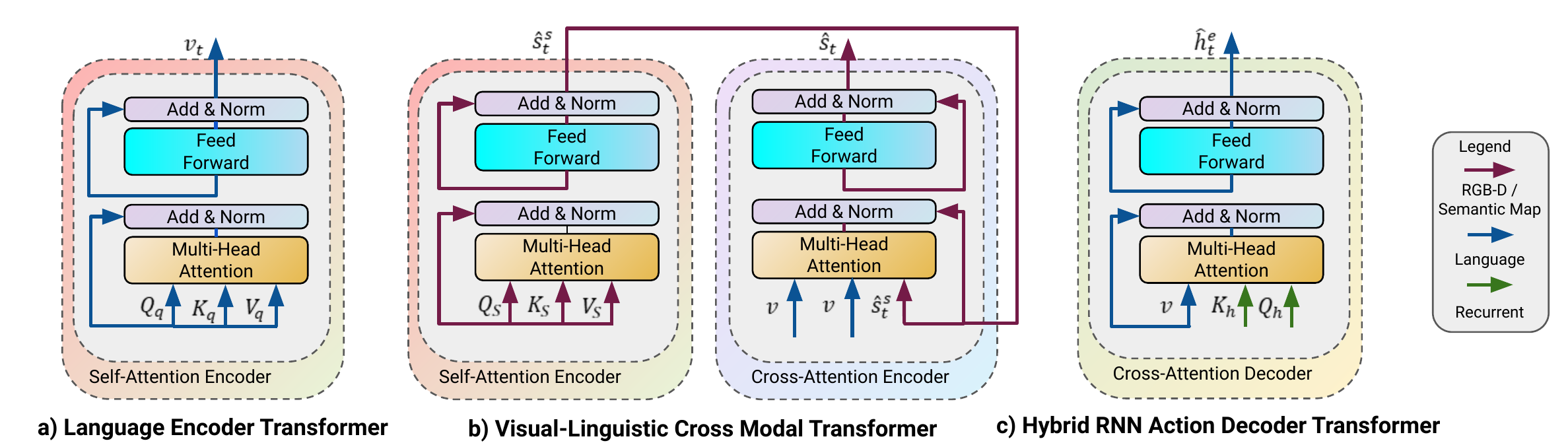}
\captionof{figure}{\textbf{Model Architecture (Individual Blocks):} Our model utilises fully-attentive Transformer blocks at each stage. \textbf{a)} We encode language using a self-attention Transformer. \textbf{b)} Visual-Linguistic attention is performed using two-stage Transformer blocks employing both self and cross-attention. \textbf{c)} Action decoder comprises of a single cross-modal Transformer module.}
  \label{SASRA3}
\end{center}
\end{figure*}

To cross-modally attend to the relevant areas of the provided instructions $q$ corresponding to the generated semantic map, we first encode these instructions $q$ in a fully attentive model as follows: 

\textbf{Language Encoder Transformer:}
Given a natural language instruction comprising of $k$ words, its representation is denoted as $\{ q_{1}, q_{2}, \ldots, q_{k} \}$ where $q_{i}$ is the embedding vector for $i_{th}$ word in the sentence. We encode the instructions using a transformer module~\cite{NIPS2017_7181} $TRM(q$ + PE$(q))$ to get the instruction encoding vector ($v$) where PE denotes the fixed sinusoidal positional encoding~\cite{NIPS2017_7181} used to encode the relative positions of words appearing in the sentence. The transformer module $TRM$ consists of stacked multi-head attention block (M.H.A) followed by a position wise feed-forward block~\cite{NIPS2017_7181} as shown in Figure \ref{SASRA3}(a) and Equation \ref{transformer}. Individual transformer blocks are computed as follows:
\begin{equation}\begin{aligned}
\text { M.H.A }(\boldsymbol{Q}, \boldsymbol{K}, \boldsymbol{V}) &=\text{concat( $\boldsymbol{h}_{1}$}, \ldots, \text{ $\boldsymbol{h}_{k}$} ) \boldsymbol{W}^{h} \\
\text {where $\boldsymbol{h}_{i}$ }&=\mathit{A}\left(\boldsymbol{Q} W_{i}^{Q}, \boldsymbol{K} W_{i}^{K}, \boldsymbol{V} W_{i}^{V}\right)
\\
\mathit{A}(\boldsymbol{Q}, \boldsymbol{K}, \boldsymbol{V})&=\operatorname{softmax}\left(\frac{\boldsymbol{Q} K^{T}}{\sqrt{d_{k}}}\right) \boldsymbol{V}
\label{transformer}
\end{aligned}\end{equation}

A basic attention block, as shown in Figure \ref{SASRA3}, uses a linear projection of the input to find queries ($Q_{q} = q$ + PE$(q) $) and keys ($K_{q} = q$ + PE$(q)$). Based on the similarity between queries ($Q_{q}$) and keys ($K_{q}$), a weighted sum over the values ($V = q$ + PE$(q)$) is computed as the output attention ($\mathit{A})$. Note that the language encoder transformer uses the same language sequences to project the tuple ($Q_{q},K_{q},V_{q}$) and hence the attention output is referred to as \textit{self-attention}. $W_{i}^{Q}, W_{i}^{K}, W_{i}^{V}$ and $W^{h}$ are parameters to be learnt. 

\textbf{Semantic-Linguistic Co-Attention Transformer:} The output representation of semantic map ($s_{t}$) is a $2r \times2r$ map centered on the agent. We refer to this map representation ($s_{t}$) as a Bag of Semantic Indices. Each cell in the map ($s_{t_{xy}}$) carries important structural information about the map. To encode the map features in a fully attentive transformer model (Equation \ref{transformer}), we construct a $2r \times2r$ relative positional encoding matrix ($\mathcal P$) using a Gaussian kernel, centered on the agent, with a scale factor of $\frac{r}{2}$. Gaussian positional encoding ($GPE$) is computed after projecting the Gaussian kernel into an embedding space as follows: 
\begin{equation}\begin{aligned}
\text{GPE}&=\it{embed_{2D}}(F(x,y)) \\
\boldsymbol{F}\left(\boldsymbol{x}, \boldsymbol{y}\right)&=\frac{b^{2}}{\sqrt{2 \pi w^{2}}} \exp\left(-\frac{\left(\boldsymbol{x}-\boldsymbol{y}\right)^{2}}{2 w^{2}}\right)
\label{GPE}
\end{aligned}\end{equation}

where $w$ is the input scale parameter (equivalent to the standard deviation of the Gaussian) and $b$ is the output scale parameter.

We encode the semantic map representation ($s_{t}$) in a 2 step process. First, we compute \textit{self-attention} over the map ($s_{t}$) by employing Equation \ref{transformer}. We use the sum of Bag of Semantic Indices and Gaussian positional encoding ($\mathcal P$) as Query, Key and Value ($Q_{s} = K_{s} = V_{s} = s_{t} + GPE(s_{t})$) in Equation \ref{transformer}. This transformer feature representation $\hat{s}_{t}^{s}$ $\in$ $\mathbb{R}^{2}$ using \textit{self-attention}, as shown in Figure \ref{SASRA3}(b), is computed as follows:
\begin{equation}
\hat{\boldsymbol{s}}_{t}^{s} = TRM(\boldsymbol{s}_{t} + GPE(\boldsymbol{s}_{t}))    
\end{equation}

$\hat{s}_{t}^{s}$ is $(r\times r)\times H$ matrix, where  $(r\times r)$ is the $2D$ semantic map collapsed into one dimension and $H$ is the hidden size. Typical values we utilised are $r=40$ and $H=512$.

Second, we perform \textit{cross-attention} between computed language features ($v$) and semantic map self attention features $\hat{s}_{t}^{s}$. We employ Equation (\ref{transformer}) by using $\hat{s}_{t}^{s} + GPE(\hat{s}_{t}^{s})$ as Query and instruction encoding ($v$) as Key and Value to get final cross-modal attention features $\hat{s}_{t}$ $\in$ $\mathbb{R}^{(r\times r)\times H}$ as follows:

\begin{equation}
\hat{\boldsymbol{s}}_{t} = TRM(\hat{\boldsymbol{s}}_{t}^{s} +GPE(\hat{\boldsymbol{s}}_{t}^{s}), \boldsymbol{v})   
\end{equation}

\subsection{RGBD-Linguistic Cross Attention Transformer}

Given an initial observed image ($r_{t}$ $\in$ $\mathbb{R}^{H_{o}\times W_{o}\times 3 }$), we generate a low-resolution spatial feature representations $f^{r}_{t}$ $\in$ $\mathbb{R}^{H_{r}\times W_{r}\times C_{r}}$  by using a classical CNN. Typical values we utilised are $H_{r}=W_{r} = 7$, $C_{r}=2048$. Furthermore, we process depth modality ($d_{t}$ $\in$ $\mathbb{R}^{H_{o}\times W_{o}}$) using a CNN pre-trained on a large scale visual navigation task i.e. DDPPO~\cite{wijmans2020ddppo} to generate a spatial feature representation $f^{d}_{t}$ $\in$ $\mathbb{R}^{H_{d}\times W_{d}\times C_{d}}$. Typical values used in DDPPO training are $H_{d}=W_{d} = 4$, $C_{d}=128$.

\textbf{Early Fusion:} We employ an early fusion of image spatial features $f^{r}_{t}$ and depth spatial features $f^{d}_{t}$. Specifically, we reduce the channel dimension of the spatial image features $f^{r}_{t}$ using a $1\times 1$ convolution and perform average adaptive pooling across spatial dimensions to reduce the dimensions from $H_{r}\times W_{r}$ to $H_{d}\times W_{d}$. We concatenate the final outputs $f^{r}_{t}$ and $f^{d}_{t}$ along the channel dimension to get a fused RGB-D representation ($i_{t}$).

\textbf{Cross Modal Attention:} We utilised Equation \ref{transformer} to perform cross modal attention between linguistic features ($v$) and fused RGB-D features ($i_{t}$). We perform this \textit{cross-attention} in two steps. First, we use the sum of $i_{t}$ and Learned Positional Encoding ($LPE(i_{t})$) as Query, Key and Value ($Q_{s} = K_{s} = V_{s} = i_{t} + LPE(i_{t})$) in Equation \ref{transformer} to compute \textit{self-attention} features $\hat{i}_{t}^{s}$ where we learn a spatial $2D$ positional encoding $LPE$ as opposed to utilising a fixed positional encoding~\cite{NIPS2017_7181}. Second, we perform \textit{cross-attention} by using $i_{t} + LPE(i_{t})$ as Query and instruction encoding ($v$) as Key and Value in Equation \ref{transformer} to get cross-modal attention features $\hat{i}_{t}$ $\in$ $\mathbb{R}^{(H_{d}\times H_{d})\times H}$

\textbf{Late Fusion:} We perform a late fusion of cross modal semantic map features $\hat{s}_{t}$ and cross modal RGB-D features $\hat{i}_{t}$. Specifically, we utilise average pooling across the spatial dimensions of $\hat{s}_{t}$ before concatenating along the hidden size dimension $H$ to get visual-linguistic embedding ($\hat{V}_{t}^{e}$)

\subsection{Hybrid Action Decoder}

For action selection, the agent preserves a temporal memory of the previous observed visual-linguistic states ($\hat{V}_{t}^{e}$) and previous actions($a_{t-1}$). We utilise a recurrent neural network $\text{RNN}$ to preserve this temporal information across time. 
\begin{equation}\boldsymbol{h}_{t}=\operatorname{RNN}\left(\left[\hat{\boldsymbol{V}}_{t}^{e}, \mathbf{a}_{t-1}\right]\right)\label{recurrence}\end{equation}

\textbf{Temporal Cross Modal Transformer:} The agent selects an action $a_{t}$ by keeping track of the completed parts of the instructions ($q$) and observed visual-linguistic states ($\hat{V}_{t}^{e}$). We preserve this temporal information regarding instruction completeness using an action decoder transformer, as shown in Figure \ref{SASRA3}(c), which performs \textit{cross-attention} between hidden states from recurrent network (Equation \ref{recurrence}) and instruction encoding ($v$). We compute this \textit{cross-attention} $TRM(\boldsymbol{h_{t}} +FPE(\boldsymbol{h_{t}}), \boldsymbol{v})$ by utilizing recurrent hidden states $h_{t}$ as Query and Key ($Q_{h} = K_{h} = h_{t} + FPE(h_{t})$) and instruction encoding ($v$) as Value in Equation \ref{transformer}. Finally, we compute the probability ($p_{t}$) of selecting the most optimal action ($a_{t}$) at each time-step by employing a feed-forward network followed by $softmax$ as follows:
\begin{equation} \begin{aligned}
\hat{\boldsymbol{h}}_{t}^{e} &= TRM(\boldsymbol{h}_{t} +FPE(\boldsymbol{h}_{t}), \boldsymbol{v}) \\
p_{t} &= \text{softmax}(\mathrm{M}(\hat{\boldsymbol{h}}_{t}^{e}))
\end{aligned}\end{equation}
where $\mathrm{M}(.)$ is a multi-layer perceptron and $TRM(.)$ is the Transformer module. \\
Conceptually, our Hybrid Action Decoder performs attention both on temporal sequences (i.e. along time) as well as language sequence. We empirically show (in Section \ref{ablationsection}) that our hybrid action decoder is indeed crucial for the long-horizon and complex task such as VLN-CE and this design choice leads to better results than a single-temporal module based on either LSTMs or Transformers alone.   

\subsection{Training}
We train our model using a cross entropy loss ($\mathcal{L}_{loss}$) computed using ground truth navigable action ($y_{t}$) and log of the predicted action probability ($p_{t}$) at step $t$. 
\begin{equation} 
\vspace{-0.1cm}
\begin{aligned}
\mathcal{L}_{loss} &= -\sum_{t}^{} y_{t}\log \left(p_{t}\right)\end{aligned}
\vspace{-0.1cm}
\end{equation}

\textbf{Training Regimes:} We explore two popular imitation learning approaches for training, i.e., 1) Teacher-Forcing \cite{williams1989learning}, 2) DAGGER \cite{ross2011reduction}. Teacher-forcing is the most popular training method for RNNs which minimizes maximum-likelihood loss using ground-truth samples \cite{williams1989learning, goyal2016professor}. However, teacher-forcing strategy suffers from exposure bias problems due to differences in training and inference \cite{ross2011reduction, krantz2020navgraph}. One popular approach to minimize the exposure bias problem in teacher-forcing method is DAGGER \cite{ross2011reduction}. In Dagger, the training data set for iteration $N$ is collected with probability $\beta^N$ (where, $0 < \beta < 1$) for ground-truth action and current policy action for remaining. The training for iteration $N$ is done using the aggregated dataset collected till iteration $N$.

\section{Experiments}
\label{sec:exp}
\textbf{Simulator and Dataset:}
We use Habitat simulator~\cite{habitat19iccv} to perform our experiments. We use VLN-CE dataset presented by Krantz et al.~\cite{krantz2020navgraph} to evaluate VLN in continuous environments. VLN-CE is built upon Matterport3D dataset~\cite{Matterport3D} which is a collection of $90$ environments captured through around 10k high-definition RGB-D panoramas. VLN-CE provides $4475$ trajectories followed by an agent inside Matterport3D simulation environment available in Habitat Simulator. Each trajectory is associated with 3 instructions annotated by humans. The corresponding dataset is divided into train, validation seen and validation unseen splits. The train set contains 10819 episodes from 61 scenes, validation seen set contains 778 episodes from 53 scenes and validation unseen set contains 1839 episodes from 11 scenes. The set of scenes in train and validation unseen splits are disjoint.

VLN-CE dataset provides the following low-level actions for each instruction-trajectory pair for navigation inside Habitat Simulator: \texttt{move forward} ($0.25m$), \texttt{turn-left} or \texttt{turn-right} ($15 deg.$) and \texttt{stop}. The trajectories in VLN-CE span $55$ steps on average, making the problem realistic and challenging to solve than previously proposed navigation-graph based R2R dataset~\cite{mattersim} which corresponds to an average trajectory length of $4-6$. Note that our experiments do not include R2R dataset~\cite{Matterport3D} since we focus on a richer and more challenging setting of VLN in continuous environments i.e. VLN-CE.

\textbf{Evaluation Metrics:} We use the following standard visual navigation metrics to evaluate our agents as described in previous works~\cite{DBLP:journals/corr/abs-1807-06757, 49206, mattersim}: Success rate ($\textbf{SR}$), Success weighted by path length ($\textbf{SPL}$), Normalized Dynamic Time Warping ($\textbf{NDTW}$), Trajectory Length ($\textbf{TL}$) and Navigation Error ($\textbf{NE}$). Following prior works \cite{Wang2019ReinforcedCM, krantz2020navgraph, DBLP:journals/corr/abs-1807-06757}, we use $\textbf{SPL}$ and $\textbf{SR}$ as the primary metrics to evaluate the navigation performance of our agent. Please see~\cite{DBLP:journals/corr/abs-1807-06757, 49206, mattersim} for details of the metrics.

\textbf{Implementation Details:} 
For Teacher-Forcing training, we train the models for at most $20$ epochs. We perform early stopping based on the performance of the model on the validation seen dataset. For DAGGER training, we start from the model pre-trained with teacher-forcing training and then fine-tune using DAGGER. We set the number of dataset aggregation rounds $N$ for DAGGER to be $3$. $\beta$ is set to $0.75$. Hence, for the $Nth$ Dagger iteration, the dataset is collected with probability $0.75^N$ for oracle policy actions blended with actions from the agent's policy obtained after training $(N-1)th$ iteration with a probability $(1-0.75^N)$. For each DAGGER round, we perform $4$ epochs of teacher forcing training. All the models are trained using three RTX 2080 Ti GPUs and Habitat-API~\cite{habitat19iccv} version 0.1.5.

To avoid loss curve plateauing, we use Adam Optimizer~\cite{kingma2014method} with a Cyclical Learning Rate~\cite{smith2015cyclical} in the range [$2^{-6}, 1^{-4}$]. The cyclic learning rate was varied over a range of epochs. The inflection weighting coefficient for imitation learning is set to $3.2$ following \cite{krantz2020navgraph}. For individual transformer blocks, we use a consistent hidden size ($H = 512$), Number of Transformer Heads ($n_{h}=4$) and size of feed-forward layer ($s_{F.F}=1024$). To improve memory efficiency, we down-scaled the generated semantic map ($S_{t}$) by a factor of $\frac{1}{2}$. The final dimension of $S_{t}$ used as an input to the Transformer module is $\frac{r}{2}\times \frac{r}{2}$ where $r=40$. We employ pre-training for vision components in our architecture. Specifically, we utilize Imagenet~\cite{7780459} pre-trained Resnet-50 for RGB and Resnet-50 pre-trained on DDPPO~\cite{wijmans2020ddppo}  to compute low-resolution features for visual modalities. We use the depth and semantic label of pixels from the simulation environment following prior works \cite{habitat19iccv, krantz2020navgraph, maast}. 

\renewcommand{\arraystretch}{0.86}
\begin{table*}[t]
\centering
\caption{\textbf{Quantitative Comparison}: Results of the proposed method, compared against several baselines, evaluated on VLN-CE dataset. Random and Forward-only are trivial non-learning baselines reported to provide context to the performance of learning based baselines. Seq2Seq and CMA are state-of-the-art models on VLN-CE dataset trained following \cite{krantz2020navgraph}. We implemented Seq2Seq-SM and CMA-SM baseline models by providing additional semantic map input to Seq2Seq and CMA models. The best results with DAGGER are in \textbf{bold} and the best results with teacher-forcing is \underline{underlined}. Following prior works \cite{Wang2019ReinforcedCM, krantz2020navgraph, DBLP:journals/corr/abs-1807-06757}, \textbf{SPL} and \textbf{SR} are considered as primary metrics for evaluation.
Please note that, as mentioned by the authors of VLN-CE~\cite{krantz2020navgraph}, different hardware + Habitat builds lead to different results\textsuperscript{$\ddagger$}. Hence, we re-trained all the models using the same build and following the implementation details presented in Sec.~\ref{sec:exp} for a fair comparison. } \label{tab:quantitative}
\resizebox{1\textwidth}{!}{
\begin{tabular}{cccc@{\hskip 0.1in}cccc@{\hskip 0.2in}ccccc}
\toprule
 & \multicolumn{2}{c}{\textbf{Training Regime}} & \multicolumn{5}{c}{\textbf{Validation Seen}} &\multicolumn{5}{c}{\textbf{Validation Unseen}} \\ 
\cmidrule(r{0.1in}){2-3} \cmidrule(r{0.2in}){4-8} \cmidrule{9-13}
Model & {Teacher-Forcing~\cite{williams1989learning}} &{DAGGER~\cite{ross2011reduction}} &  \textbf{SR}~$\uparrow$ & \textbf{SPL}~$\uparrow$ & \textbf{NDTW}~$\uparrow$ & \textbf{TL}~$\downarrow$ & \textbf{NE}~$\downarrow$ & \textbf{SR}~$\uparrow$ & \textbf{SPL} & \textbf{NDTW}~$\uparrow$ & \textbf{TL}~$\downarrow$ & \textbf{NE}~$\downarrow$ \\
\midrule
Random & & &  0.02 & 0.02 & 0.28 & 3.54 & 10.20 & 0.03& 0.02 & 0.30 & 3.74 & 9.51 \\
Forward-Only & & &  0.04 & 0.04 & 0.33 & 3.83 & 9.56 & 0.03& 0.02 & 0.30 & 3.71 & 10.34 \\
\midrule
\multirow{2}{1.8cm}{\centering Seq2Seq} &\checkmark & &  0.25 & 0.24 & 0.47 & 8.15 & 8.49 & 0.17& 0.16 & 0.43 & 7.55 & 8.91  \\
 & &\checkmark & 0.32 & 0.31 & 0.51 & 8.49 & 7.61 & 0.19 &  0.17 & 0.43 & 7.78 & 8.69  \\
 \midrule
\multirow{2}{1.9cm}{\centering Seq2Seq-SM} &\checkmark& & 0.26& \underline{0.25} & \underline{0.48} & \underline{7.92} & 8.13 &  0.16 &  0.15 &  0.41 &  7.54 &  9.14 \\
 & & \checkmark & 0.30 &  0.28 & 0.49 &  8.58 &  7.98 & 0.19 & 0.17 & 0.43 & 7.69 & 8.98 \\
 \midrule
\multirow{2}{1.8cm}{\centering CMA} &\checkmark&  & 0.24 &  0.23 & 0.47 & 7.94 & 8.69 & 0.19 & 0.18 & \underline{0.44} & \underline{6.96} & \underline{8.52}  \\
 && \checkmark & 0.28 & 0.26 & 0.49 & \textbf{8.43} & 8.12 & 0.20 & 0.18 & 0.44 & \textbf{7.67} & 8.72 \\
\midrule
\multirow{2}{1.8cm}{\centering CMA-SM} &\checkmark& & \underline{0.27} & \underline{0.25} & \underline{0.48} & 8.44 &  8.25 & 0.18	& 0.16 & \underline{0.44} &	7.52 & 8.73 \\
 && \checkmark & 0.29 & 0.28 & 0.49 & 8.49 & 7.79 & 0.19 & 0.18 & 0.43 & 7.70 & 8.92 \\
\midrule
\multirow{2}{1.8cm}{\centering \textbf{SASRA (Proposed)}} &\checkmark& & \underline{0.27} &	\underline{0.25} &	0.46	&8.84 &	8.48&	\underline{0.22}&	\underline{0.21}&	\underline{0.44} &	8.05&	8.75  \\
 &&\checkmark & \textbf{0.36} &	\textbf{0.34}&	\textbf{0.53}	&8.89 &	\textbf{7.17} & \textbf{0.24}	& \textbf{0.22}	& \textbf{0.47}	& 7.89	& \textbf{8.32} \\
\bottomrule
\end{tabular}}
\vspace{-0.2cm}
\end{table*}
\renewcommand{\thefootnote}{\fnsymbol{footnote}}
\footnotetext[3]{\tiny \url{https://github.com/jacobkrantz/VLN-CE\#published-results-vs-leaderboard}}

\subsection{Quantitative Comparison with Baselines}
We report the quantitative results on VLN-CE dataset in Table~\ref{tab:quantitative}. We implement several baselines to compare the performance of the proposed model. The compared baseline models are: \textbf{1) Blind (i.e., Random and Forward-Only):} These agents does not process any sensor input and have no learned component \cite{krantz2020navgraph}. The Random agent randomly performs an action based on the action distribution of train set. The Forward-Only agent starts with a random heading and takes a fixed number of actions. The performance of these agents are reported from \cite{krantz2020navgraph}. \textbf{2) Sequence to Sequence (Seq2Seq):} Seq2Seq agent employs a recurrent policy that takes a representation of instructions and visual observation (RGB and Depth) at each time step and then predicts an action (\cite{mattersim, krantz2020navgraph}). \textbf{3) Sequence to Sequence with Semantics (Seq2Seq-SM):} Seq2Seq model which includes additional visual observation, (i.e., semantic map) to aid learning. \textbf{4) Cross Modal Matching (CMA):} A sequence to sequence model with cross-modal attention and spatial visual reasoning (\cite{Wang2019ReinforcedCM, krantz2020navgraph}). \textbf{5) CMA with Semantics (CMA-SM):} CMA Model which includes additional visual observation, (i.e., semantic map) to aid learning. 

The Seq2Seq and CMA are state-of-the-art baseline models in VLN-CE implemented following \cite{krantz2020navgraph}. Seq2Seq-SM and CMA-SM are our implemented baselines where base Seq2Seq and CMA models are enhanced by providing semantic map self-attention representation as additional input. The performance of the all the baselines are reported in Table~\ref{tab:quantitative} with both teacher-forcing and DAGGER training. We report both Validation Seen (val-seen) and Validation Unseen (val-unseen) sets. 

\underline{State-of-the-art Baselines.} It is evident from Table~\ref{tab:quantitative} that the proposed SASRA agent achieves significant performance improvement over state-of-the-art VLN-CE baseline models, i.e., Seq2Seq and CMA. We observe that Seq2Seq and CMA show overall comparable performance. Seq2Seq performs slightly better in val-seen set than CMA, however, CMA performs slightly better in val-unseen. On the other hand, the proposed SASRA agent shows performance improvement compared to both models in both validation sets and with both training regimes. With DAGGER training, we find that the improvement in SPL is $+5\%$ absolute ($+29\%$ relative) in val-unseen and $+3\%$ absolute ($+9.6\%$ relative) in val-seen compared to Seq2Seq. With teacher-forcing training, the performance improvement in SPL is $+5\%$ absolute ($+31\%$ relative) in val-unseen and +1\% absolute ($+4\%$ relative) in val-seen compared to Seq2Seq. Compared to CMA agent with DAGGER training, the absolute performance improvement in SPL is $+4\%$ in val-unseen and $+8\%$ in val-seen. With teacher-forcing training, the absolute performance improvement in SPL is $+3\%$ in val-unseen and $+2\%$ in val-seen. In terms of SR, we also see similar performance improvement.  

\underline{Baselines with Semantic Map.} From Table~\ref{tab:quantitative}, we observe that the Seq2Seq-SM model despite having access to semantic maps is not able to consistently perform better compared to Seq2Seq. We find a performance drop for Seq2Seq-SM compared to Seq2Seq in val-unseen with teacher-forcing training (SPL of $0.15$ for Seq2Seq-SM vs  $0.16$ for Seq2Seq) and in val-seen set with DAGGER training (SPL of $0.28$ for Seq2Seq-SM vs  $0.31$ for Seq2Seq). We also do not see a consistent performance improvement across different settings with CMA-SM compared to the CMA model. CMA-SM shows better performance in val-seen set but the performance is slightly lower than CMA in val-unseen set. We think that Seq2Seq-SM and CMA-SM models are not able to effectively relate semantic map and language information needed for improved navigation performance. We expect that additional semantic map cues would lead to significant improvement in VLN performance. However, an ineffective use might lead to ambiguity and degraded performance as evident from the performance of Seq2Seq-SM and CMA-SM. This inspires the design of our proposed SASRA model which focuses on effective integration of semantic mapping and language cues. We later show in the ablation study (Table~\ref{tab:ablation}) that addition of semantic map cues results in huge performance improvement for the proposed SASRA model. Among the models utilizing semantic map cues in Table~\ref{tab:quantitative}, We observe that Seq2Seq-SM and CMA-SM show comparable performance to proposed model in some metrics, but the proposed model shows significant overall performance improvement (e.g., absolute improvement of $+6\%$ SPL in val-seen and $+5\%$ SPL in val-unseen compared to both Seq2Seq-SM and CMA-SM). 

\underline{Training Regimes.} We observe from Table~\ref{tab:quantitative} that the models trained with DAGGER in general performs better than the models trained with teacher-forcing which is expected. We find DAGGER training to be most beneficial for our SASRA agent and performance improves significantly (e.g., SPL of $0.25$ with teacher-forcing vs $0.34$ with DAGGER).  With teacher-forcing, we find several baselines showing similar performance to the proposed SASRA agent in val-seen, however, the proposed agent performs significantly better in val-unseen. With DAGGER training, we see that the proposed model shows consistent large performance improvement over all comparing baselines across val-seen and val-unseen sets. Hence, the proposed model not only performs better in previously explored environments but also generalizes better to unseen environments.

\renewcommand{\arraystretch}{0.96}
\begin{table}[t]
\scriptsize
\centering
\caption{\textbf{Ablation Study}: Study of the Proposed SASRA Agent on VLN-CE validation-seen set to investigate the impact of different components (i.e., Semantic Map (SM), Hybrid Action Decoder (HAD) and DAGGER (DA) training) in performance.}
\vspace{-0.15cm}
\label{tab:ablation}
\resizebox{0.48\textwidth}{!}{%
\begin{tabular}{c@{\hskip 0.1in}cccccccc} \\
\toprule
& & & & \multicolumn{5}{c}{\textbf{Metrics}} \\ 
\cmidrule(){5-9}
 {\textbf{\#}} & \text{SM} &\text{HAD} &\text{DA} & \textbf{SR}~$\uparrow$ & \textbf{SPL}~$\uparrow$ & \textbf{NDTW}~$\uparrow$ & \textbf{TL}~$\downarrow$  & \textbf{NE}~$\downarrow$  \\
\midrule
  1 & \checkmark & & & 0.22 & 0.20 & 0.41 & 11.07 & 9.09 \\
  2 && \checkmark &  & 0.23 & 0.22 & 0.42 & 8.30 & 8.78  \\
 3 &\checkmark& \checkmark&  & 0.27 & 0.25 & 0.46 & 8.84 & 8.48 \\
 4 &\checkmark&\checkmark &\checkmark & 0.36 & 0.34 & 0.54 & 9.19 & 7.03 \\
\bottomrule
\end{tabular}}
\vspace{-0.15cm}
\end{table}

\begin{figure*}[htp]
\centering
\includegraphics[width=0.95\textwidth]{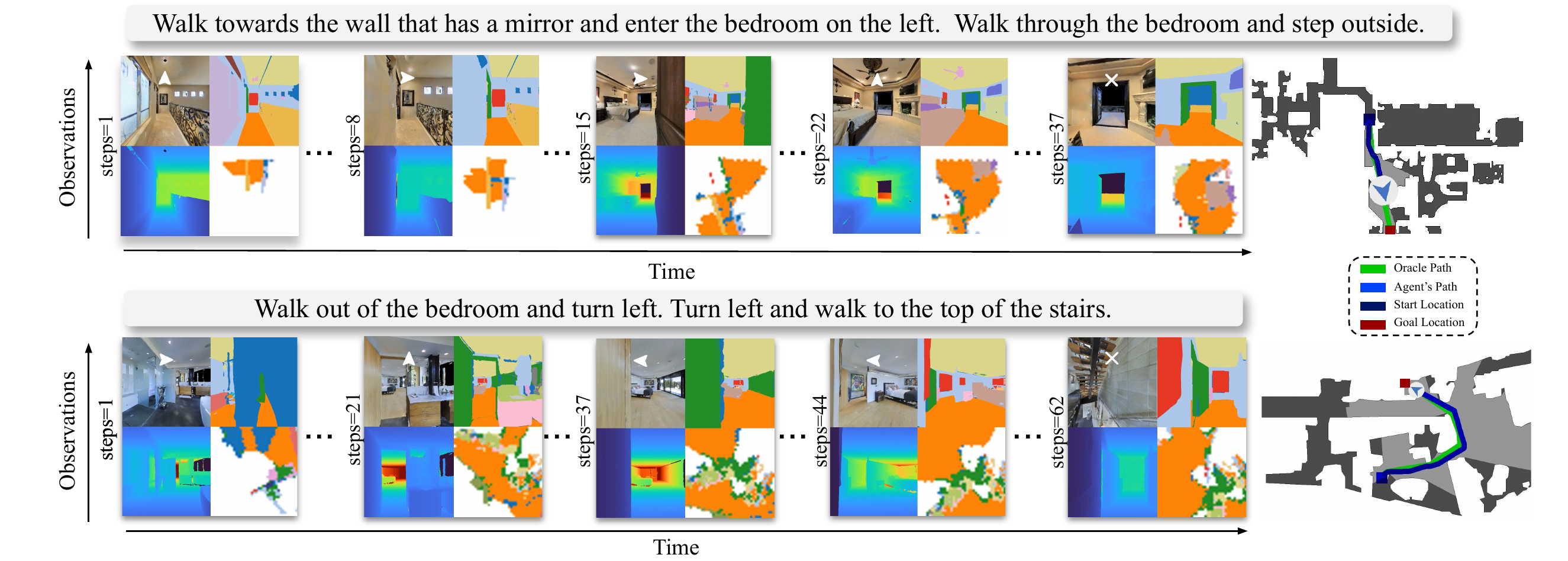}
\captionof{figure}{
\textbf{Qualitative Analysis:} Figure shows instruction-following trajectories of our SASRA agent in unseen environments within VLN-CE. Sample observations (Clockwise: RGB, Semantics, Depth and Semantic Map) seen by the agent and the corresponding actions (overlayed with RGB) are shown at each timestep. Note that the top-down map (shown on the right) is not available to the agent and is only used for performance evaluation.
}
\label{qualitative}
\vspace{-0.2cm}
\end{figure*}

\subsection{Ablation Study}
\label{ablationsection}

We perform an ablation study in Table~\ref{tab:ablation}
to investigate the impact of different components of our SASRA agent. We divide the Table into 4 rows to aid our study. 
First, we observe the effectiveness of our of semantic map attention (i.e, SLAM-T) module comparing row 2 and row 3. We see that use of semantic map cues (row 3) leads to significant performance gain (i.e., $+4\%$ absolute in SR and $+3\%$ absolute in SPL) for our SASRA agent. Second, comparing row 1 and row 3, we evaluate the effect of our Hybrid Action Decoder (HAD) module. The agent in row 1 utilizes a simple RNN module as action decoder (similar to Seq2Seq and CMA baselines) instead of our HAD module. We find that HAD module (row 3) leads to large performance gains (i.e., $+5\%$ absolute in both SR and SPL) compared to the agent without HAD module (row 1). This indicates that effective combination of semantic map attention and action decoder module is crucial for the performance of SASRA agent. Third, we evaluate the effect of DAGGER training on SASRA agent comparing row 3 and row 4. We observe that the proposed SASRA agent (row 4) with DAGGER training achieves huge improvement compared to SASRA agent with teacher forcing training (row 3). The improvement is $+9\%$ absolute in both SR and SPL (Table~\ref{tab:ablation}). This shows the importance of addressing the exposure bias issue for VLN. 

\subsection{Qualitative Results}

We qualitatively report the performance of SASRA agent in unseen photo-realistic simulation environments. Figure \ref{qualitative} shows the agent following a complex sequence of instructions in an unseen environment during training. The agent builds a top-down spatial memory in the form of semantic map and aligns the language and map features to complete the long-horizon navigation task in 37 and 62 steps. This task is significantly longer and more complex than earlier proposed R2R VLN setting~\cite{mattersim} (around 4-6 steps on average). Note that the top-down map (shown on the right) is not available to the agent during training or inference and is only used for performance evaluation.

In Figure 4, Episode 1 represents a failure case for our agent. We see that our agent is able to reach quite close to the goal for this episode but stops before reaching success radius and the episode is registered as failure. We see that
agent follows instructions to enter the bedroom and walk through. However, it is confused by the phrase ``step outside” which is a bit ambiguous as there are no details to indicate which way (as there are multiple doors) or where it needs to step outside. Episode 2 represents a success case for our agent. We observe that our agent is able to follow the instructions correctly and reach its goal. Although there are multiple instructions of ``turn left", the agent is able to turn left with the correct number of steps to effectively reach its goal. In this environment, there is a staircase going up and down and the goal location is close to that. The agent is able to immediately understand that it is already at top of stairs in the current floor and does not need to walk up the stairs. We believe semantic cues from environment helps the agent significantly in this case.

\section{Conclusions}

In this paper, we propose an end-to-end learning-based semantically-aware navigation agent to address the Vision-and-Language Navigation task in continuous environments. Our proposed model, SASRA, combines classical semantic map generation in an attentive transformer-like architecture to focus on and extract cross-modal features between local semantics map and provided instructions for effective VLN performance in a new environment. Extensive experiments and ablation study illustrate the effectiveness of the proposed approach in navigating both previously explored environments and unseen environments. The proposed SASRA model achieves relative performance improvement of over $22\%$ in unseen validation set compared to the baselines.

{\small
\bibliographystyle{ieee_fullname}
\bibliography{bibliography}
}

\end{document}